**Transformer-Based Classification of Bacterial Raman Spectra with LOOCV**

Jamile Mohammad Jafari[a,b], Thomas Bocklitz[a,b*]

[a] Leibniz Institute of Photonic Technology, Member of Leibniz Health Technologies, Member of the Leibniz Centre for Photonics in Infection Research (LPI), Albert-Einstein-Strasse 9, 07745, Jena, Germany

[b] Institute of Physical Chemistry (IPC) and Abbe Center of Photonics (ACP), Friedrich Schiller University Jena, Member of the Leibniz Centre for Photonics in Infection Research (LPI), Helmholtzweg 4, 07743, Jena, Germany

*Thomas.bocklitz@uni-jena.de

Abstract

Transformer-based models have recently attracted increasing attention for Raman spectral classification. In this study, a transformer-based approach was systematically evaluated using a nested leave-one-replicate-out cross-validation framework and compared with conventional machine-learning pipelines combining PCA or ICA with LDA, SVM, and Random Forest classifiers. A bacterial Raman dataset comprising 5,417 single-cell spectra from six bacterial species and nine independent measurement replicates was used.

The transformer consistently achieved the highest classification performance across independent test replicates and significantly outperformed all conventional approaches. Analysis of the learned latent feature space revealed improved class separation compared with PCA- and ICA-based representations. Furthermore, the transformer maintained superior performance when applied directly to raw Raman spectra without preprocessing, demonstrating robust behavior across measurement replicates.

These findings highlight the potential of transformer-based models for robust Raman spectral classification and emphasize the importance of replicate-aware validation for realistic model evaluation.

Keywords: Transformer model; Raman spectroscopy; Chemometrics; Classical Machine learning; Cross-validation

## 1 Introduction

Raman spectroscopy is a non-destructive analytical technique that provides detailed information about the chemical composition and molecular structure of materials [1, 2]. The technique has been widely applied in biomedical research, including toxicology [3], forensics [4], drug discovery [5, 6], metabolic studies [7], and in vivo diagnostics [8], as well as in pharmaceutical analysis [9] and microbiology [10], due to its ability to generate label-free molecular fingerprints.

Despite the rich molecular information contained in Raman spectra, their interpretation remains challenging. Raman datasets are often characterized by high dimensionality, spectral redundancy, and subtle differences between classes, making reliable discrimination difficult through visual inspection alone [11, 12]. To address these challenges, chemometric methods have become an essential component of Raman data analysis. Techniques such as principal component analysis (PCA) and partial least squares (PLS) have been widely employed for dimensionality reduction and feature extraction, while classifiers including linear discriminant analysis (LDA), support vector machines (SVM), and random forests (RF) have been widely employed for dimensionality reduction, feature extraction, and classification of Raman spectra [13-17]. More recently, machine learning approaches have further expanded the analytical capabilities of Raman spectroscopy, enabling automated pattern recognition and classification across a wide range of applications [18].

Although conventional chemometric and machine learning methods have achieved considerable success in Raman spectral analysis, their performance may be limited when complex, nonlinear relationships are present in the data. Recent advances in deep learning have provided new opportunities for spectral analysis by enabling models to learn informative representations directly from raw spectral data [19]. Consequently, deep learning approaches have attracted increasing attention for Raman spectral classification and related analytical tasks, including bacterial identification, disease diagnostics, and chemical analysis [20-23].

Among modern deep learning architectures, transformer models have recently emerged as a promising alternative for spectral analysis. Originally developed for natural language processing, transformers employ self-attention mechanisms that enable the modeling of long-range dependencies within sequential data [24-26]. Since Raman spectra can be regarded as one-dimensional sequential signals, transformer architectures may offer advantages in capturing spectral relationships distributed across multiple Raman shifts. Several recent studies have reported encouraging results using transformer-based models for Raman spectroscopy applications [27-31].

Despite these promising findings, most existing studies have focused primarily on developing new transformer architectures and achieving high classification performance on specific datasets. Although comparisons with conventional machine-learning approaches are often reported, model evaluation is frequently based on random train-test splits or standard cross-validation strategies. In Raman spectroscopy, such evaluation schemes may lead to overly optimistic performance estimates because spectra acquired from the same measurement replicate can appear in both the training and test sets [32].

In practical applications, however, classification models are expected to generalize to spectra acquired from previously unseen measurement replicates. Variations in experimental conditions, instrument performance, sample preparation, and measurement sessions can introduce substantial replicate effects that significantly influence model performance. Therefore, replicate-wise validation strategies provide a more realistic assessment of model robustness and generalization capability. Despite their practical importance, systematic evaluations of transformer-based Raman classification under rigorous replicate-aware validation frameworks remain limited.

Therefore, the objective of this study was to systematically evaluate the performance of a transformer-based approach for Raman spectral classification under strict replicate-wise validation conditions. Using a bacterial Raman spectroscopy dataset and a nested leave-one-replicate-out cross-validation strategy, we compared the performance of the transformer model with widely used conventional approaches, including PCA- and ICA-based classification pipelines combined with LDA, SVM, and RF classifiers. Particular emphasis was placed on evaluating model performance on independent replicates of measurements and determining whether transformer-based models can maintain competitive classification performance when applied to both preprocessed and raw Raman spectra.

## 2. Experimental and methods

### 2.1. Datasets

#### 2.1.1. Bacterial Raman Dataset

A previously published dataset of bacterial Raman spectra was used in this study [33]. The dataset contains single-cell Raman spectra acquired from six bacterial species: *Escherichia coli* DSM 423, *Klebsiella terrigena* DSM 2687, *Pseudomonas stutzeri* DSM 5190, *Listeria innocua* DSM 20649, *Staphylococcus warneri* DSM 20316, and *Staphylococcus cohnii* DSM 20261. For each species, spectra were collected across nine independent measurement replicates, yielding a total of 54 species-replicate combinations.

After preprocessing, the dataset comprised 5,417 single-cell Raman spectra, with each spectrum represented by 584 Raman shift variables. The sample sizes for each bacterial species are summarized in Table 1. The use of multiple independent replicates makes this dataset particularly suitable for evaluating model robustness and generalization under replicate-wise validation schemes. Additional details regarding bacterial cultivation, spectral acquisition, and preprocessing procedures are provided in the original publication [33]. Mean Raman spectra of the six bacterial species are shown in Fig. 1.

Table 1: Sample size of the bacteria dataset. The information was summarized based on preprocessed spectra

| **Species** | ***Escherichia coli*** | ***Pseudomonas stutzeri*** | ***Klebsiella terrigena*** | ***Listeria innocua*** | ***Staphylococcus cohnii*** | ***Staphylococcus warneri*** |
|---|---|---|---|---|---|---|
| #Spectra | 894 | 902 | 910 | 916 | 898 | 897 |

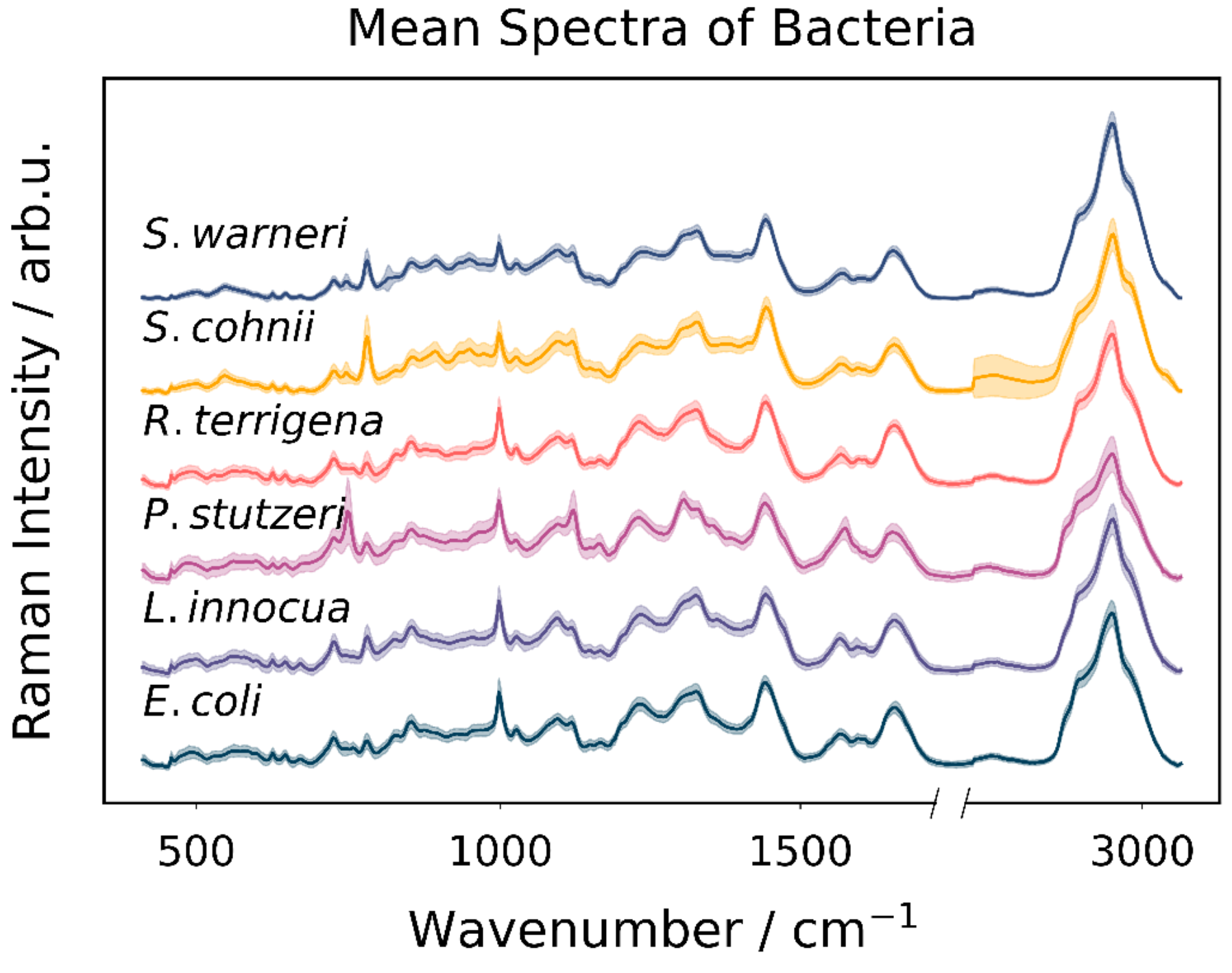

Fig.1. The mean spectra of the Raman spectra dataset were obtained from six distinct bacterial species.

### 2.2. Data analysis

The analysis of Raman spectral data followed a structured workflow, as illustrated in Fig. 2. Raw Raman spectra were first preprocessed to remove noise, calibrate wavenumbers, correct baselines, and apply vector normalization. Following spectral preprocessing, Raman spectra were analyzed using two complementary strategies. In the conventional machine learning workflow, feature extraction was first performed using Principal Component Analysis (PCA) or Independent Component Analysis (ICA), followed by classification with Linear Discriminant Analysis (LDA), Support Vector Machines (SVMs), or Random Forests (RFs). In parallel, a transformer-based model was applied directly to the preprocessed spectra, without an explicit feature-extraction step. The performance of all approaches was evaluated using a nested leave-one-replicate-out cross-validation framework. Detailed descriptions of the individual models and validation procedures are provided in the following sections.

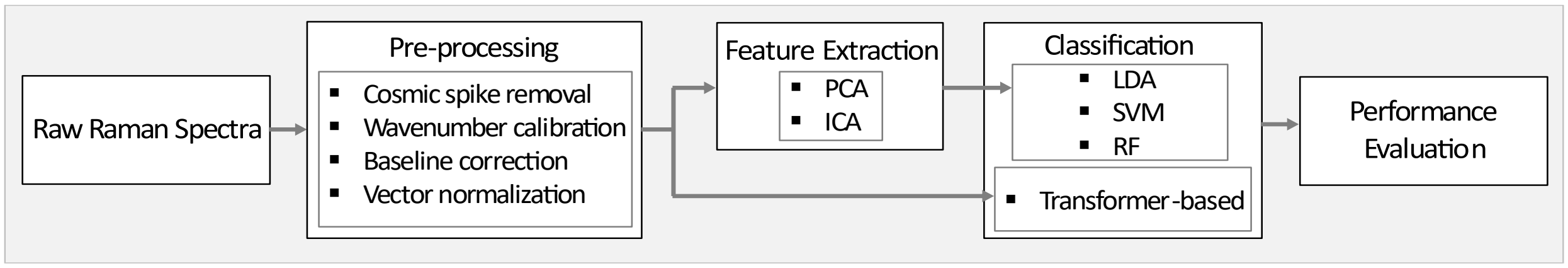

Fig.2. Overall workflow for bacterial Raman spectral classification. Following preprocessing, conventional machine learning approaches combined PCA or ICA with LDA, SVM, and Random Forest classifiers, whereas the transformer-based model operated directly on the preprocessed spectra. All approaches were evaluated using the same nested leave-one-replicate-out cross-validation framework

#### 2.2.1. Conventional Classification Models

Principal Component Analysis (PCA) and Independent Component Analysis (ICA) were used as feature extraction techniques to reduce the dimensionality of the Raman spectra.

PCA identifies orthogonal directions that capture the maximum variance in the data, whereas ICA seeks statistically independent latent components.

The extracted features were subsequently used as inputs for Linear Discriminant Analysis (LDA), Support Vector Machines (SVM), and Random Forest (RF) classifiers. LDA is a supervised linear classifier that maximizes class separability by increasing between-class variance while reducing within-class variance. SVM constructs an optimal decision boundary by maximizing the margin between classes in the feature space. RF is an ensemble learning method that combines multiple decision trees to improve classification robustness and capture nonlinear relationships within the data.

The combination of feature extraction and classification methods yielded six conventional classification pipelines: PCA-LDA, ICA-LDA, PCA-SVM, ICA-SVM, PCA-RF, and ICA-RF. These models served as benchmark approaches for comparison with the transformer-based model.

#### 2.2.2. Transformer Model

The transformer architecture employed in this study is illustrated in Fig. 3. Transformers are deep learning models designed to process sequential data via attention mechanisms, enabling modeling of both local and long-range dependencies within a sequence. Unlike conventional machine learning approaches, which separate feature extraction and classification into distinct steps, transformer models learn feature representations and perform classification within a unified end-to-end framework.

As shown in Fig. 3, each preprocessed Raman spectrum was first partitioned into a sequence of spectral patches via tokenization. The resulting tokens were subsequently mapped into a latent feature space using a linear projection layer. A learnable classification token (CLS) was prepended to the token sequence to aggregate information from all spectral patches and provide a global representation of the input spectrum. Positional embeddings were further incorporated to preserve the sequential ordering of the spectral information.

The embedded token sequence was then processed by a stack of transformer encoder blocks. Each encoder block consisted of two principal components: a multi-head self-attention module and a feed-forward neural network. The self-attention mechanism enables the model to learn relationships between different spectral regions by assigning adaptive importance weights to individual tokens, while the feed-forward network further transforms the learned representations into a higher-level feature space. Layer normalization and residual connections were incorporated throughout the encoder architecture to facilitate stable training and efficient information propagation.

Following the encoder stack, the final representation of the CLS token was extracted and forwarded to a multilayer perceptron (MLP) classification head. The MLP transformed the learned spectral representation into class probabilities corresponding to the six bacterial species, thereby generating the final classification output.

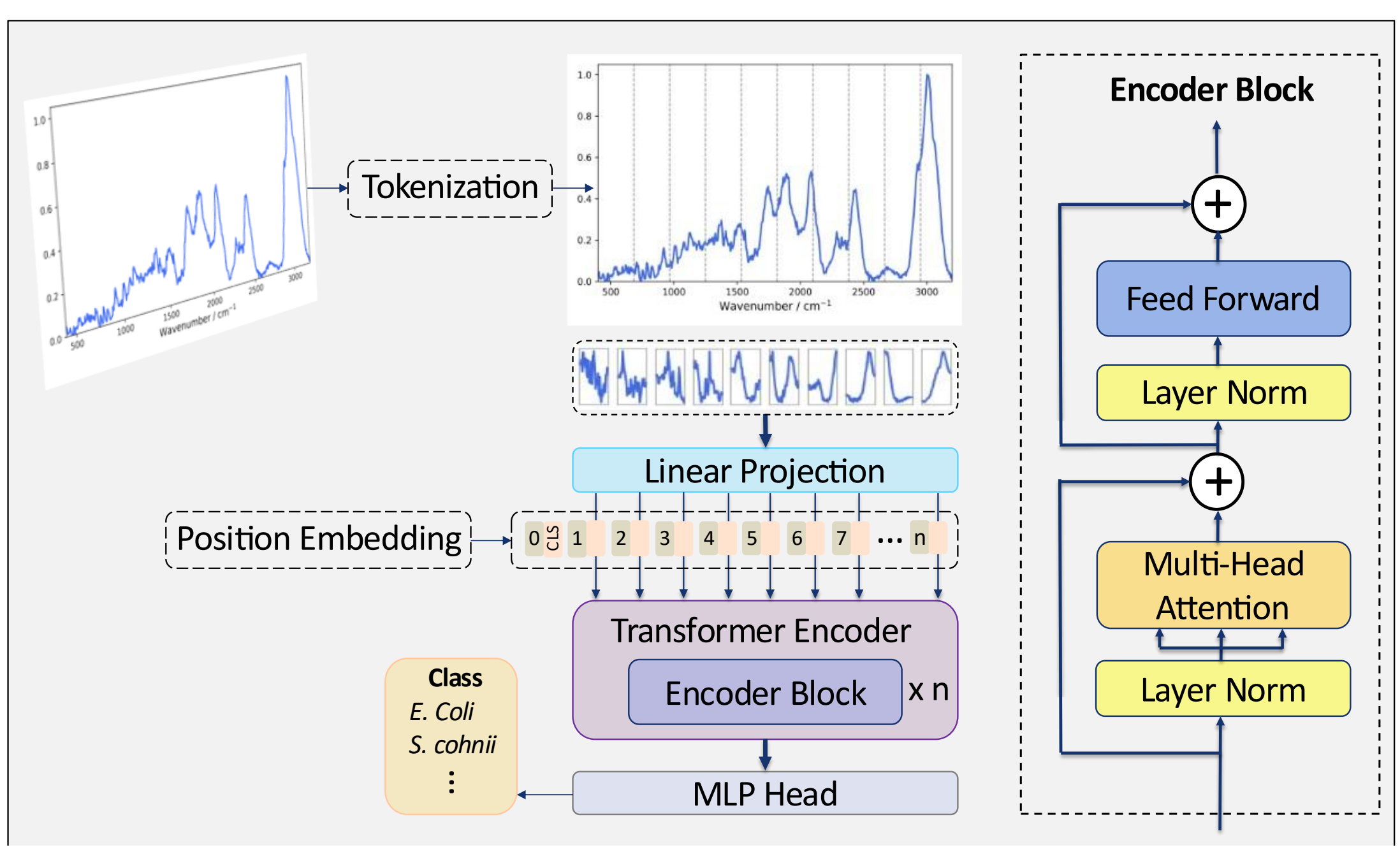

Fig. 3. Architecture of the transformer model used for Raman spectral classification.

### 2.2.3. Performance Evaluation

The performance of all classification models was evaluated using a nested leave-one-replicate-out cross-validation strategy, as illustrated in Fig. 4 [32]. The bacterial dataset consisted of nine independent measurement replicates per species. In the external cross-validation loop, one replicate was iteratively held out as an independent test set, while the remaining eight replicates were used for model development. Within each external fold, hyperparameter optimization was performed using an internal leave-one-out cross-validation on the training data. After selecting the optimal hyperparameters, the final model was trained using all spectra from the eight training replicates and subsequently evaluated on the unseen test replicate. This procedure was repeated until each replicate had served once as the independent test set. The overall classification performance was quantified using mean sensitivity, calculated from predictions on the external test sets, as defined in Equation (1) for an $l$-group classification task:

$$\bar{S} = {}^{1}/_{l} \sum_{i=1}^{l} s(i) \tag{1}$$

To assess statistical differences between classification methods, pairwise Wilcoxon signed-rank tests were performed using the replicate-wise mean sensitivity values obtained from the nine external test replicates. Statistical significance was evaluated at a significance level of $p < 0.05$.

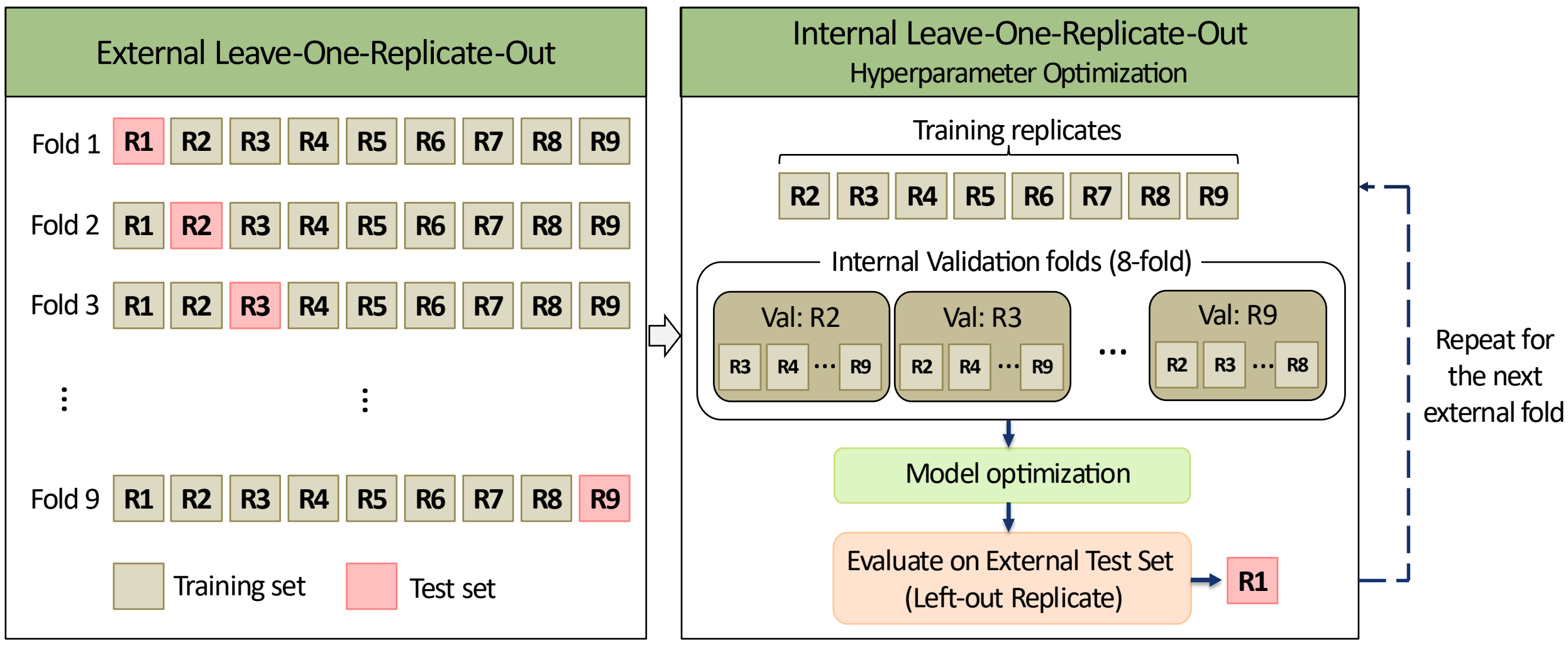

Fig. 4. Workflow of the nested leave-one-replicate-out cross-validation framework used for model optimization and performance evaluation. During external validation, one replicate was held out as an independent test set, while the remaining replicates were used for model development. Model optimization was performed using an internal leave-one-replicate-out cross-validation procedure on the training replicates. The optimized model was subsequently evaluated on the left-out test replicate. The same validation framework was applied to all classification methods.

## 3. Results and discussion

This section evaluates the performance of transformer-based and conventional machine-learning approaches for bacterial Raman spectral classification. The methods investigated include PCA-LDA, PCA-SVM, ICA-LDA, ICA-SVM, PCA-RF, ICA-RF, and a transformer-based classifier. Model performance was assessed using a nested leave-one-replicate-out cross-validation framework to provide a realistic estimate of generalization to unseen measurement replicates. First, classification results obtained from preprocessed Raman spectra are compared, and both overall and replicate-wise performance are analyzed. Subsequently, the same evaluation is performed on raw spectra to assess whether the transformer can maintain competitive classification performance without extensive spectral preprocessing.

### 3.1. Classification Performance Using Preprocessed Raman Spectra

The classification performance of the transformer-based model and conventional machine-learning approaches was first evaluated using preprocessed Raman spectra. Conventional classification pipelines combined PCA or ICA feature extraction with LDA, SVM, and Random Forest classifiers, while the transformer model was trained directly on the spectral data. For the PCA- and ICA-based models, the optimal number of components was determined during the inner cross-validation procedure using the maximum mean sensitivity criterion. All models were evaluated using the same nested leave-one-replicate-out cross-validation framework. Figure 5 summarizes the resulting classification performance across the nine independent measurement replicates: Fig. 5A presents the overall distribution of mean sensitivity values, and Fig. 5B illustrates the replicate-wise performance of each method.

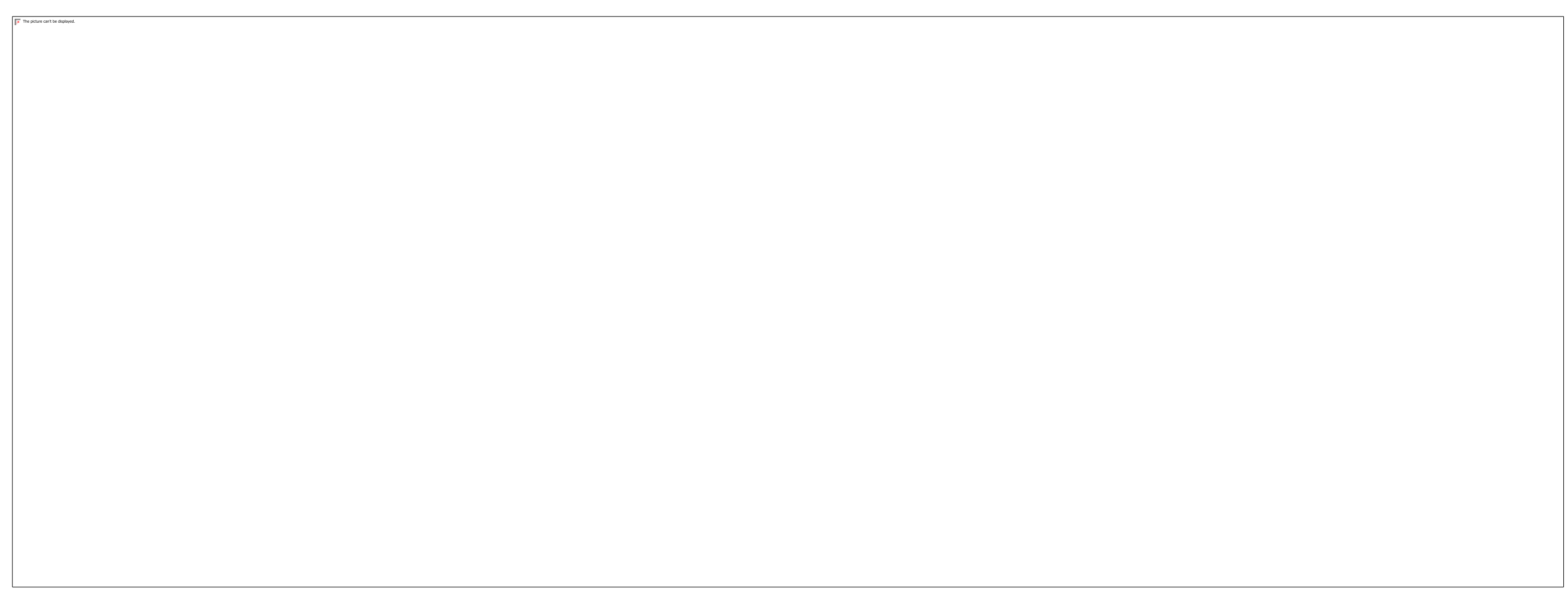

Fig. 5. Classification performance across the nine external test replicates. (A) Boxplots of mean sensitivity. (B) Heatmap of replicate-wise mean sensitivity.

As shown in Fig. 5A, the transformer model achieved the highest overall classification performance among all evaluated methods, exhibiting the highest median sensitivity across the nine measurement replicates. The PCA- and ICA-based approaches, combined with LDA or SVM, showed comparable performance and relatively similar sensitivity distributions. In contrast, the Random Forest-based models yielded lower overall performance and greater variability between replicates, particularly for ICA-RF. Overall, the transformer consistently outperformed the conventional machine-learning approaches and demonstrated stable performance across the independent measurement replicates.

The replicate-wise classification results are shown in Fig. 5B. While classification performance varied across measurement replicates, consistent trends were observed for all evaluated methods. Replicates 3 and 6 generally yielded the highest sensitivity values, whereas Replicate 7 showed lower performance across most classification models. Despite these replicate-dependent variations, the transformer consistently achieved the highest or joint-highest sensitivity values across nearly all replicates. Its advantage was particularly evident in Replicates 1, 4, 5, and 8, where noticeably higher sensitivities were observed compared with conventional machine-learning approaches. These replicate-wise results highlight the importance of evaluating classification models on independent measurement replicates, as substantial performance differences were observed between individual replicates.

To further assess the observed performance differences, pairwise comparisons between the transformer and the conventional machine-learning approaches were performed using the Wilcoxon signed-rank test based on the replicate-wise sensitivity values. The transformer significantly outperformed all conventional machine-learning models, with p-values ranging from 0.0020 to 0.0248 (Table 2). These findings confirm that the transformer's superior performance over conventional approaches was statistically significant across evaluation replicates.

Table 2. Wilcoxon Test p-values for pairwise comparisons of different models

| **Models** | **Wilcoxon p-values*** |
|---|---|
| PCA-LDA | 0.0248 |
| ICA-LDA | 0.0173 |
| PCA-SVM | 0.0020 |
| ICA-SVM | 0.0057 |
| PCA-RF | 0.0020 |
| ICA-RF | 0.0058 |

* p-values obtained from pairwise Wilcoxon signed-rank tests comparing each model against the transformer classifier.

Fig. 6 presents the confusion matrix obtained from the combined predictions of all external test replicates, providing insight into class-wise classification performance and remaining misclassifications. Overall, the transformer model achieved high classification accuracy across all six bacterial species, as indicated by the strong diagonal dominance in the confusion matrix. The highest sensitivity was observed for *P. stutzeri* DSM 5190 (97%), followed by *S. warneri* DSM 20316 (91%) and *S. cohnii* DSM 20261 (88%). Slightly lower sensitivities were obtained for *E. coli* DSM 423 (83%), *R. terrigena* DSM 2687 (81%), and *L. innocua* DSM 20649 (79%).

The most prominent misclassification occurred between *E. coli* DSM 423 and *R. terrigena* DSM 2687, with 8% of E. coli spectra misclassified as R. terrigena and 11% of R. terrigena spectra misassigned to *E. coli*. In addition, *L. innocua* DSM 20649 showed noticeable confusion with *S. warneri* DSM 20316, with 13% of the spectra being classified as *S. warneri*. Despite these remaining classification errors, the overall pattern demonstrates that the transformer learned highly discriminative representations of the bacterial Raman spectra, resulting in robust classification performance across all species.

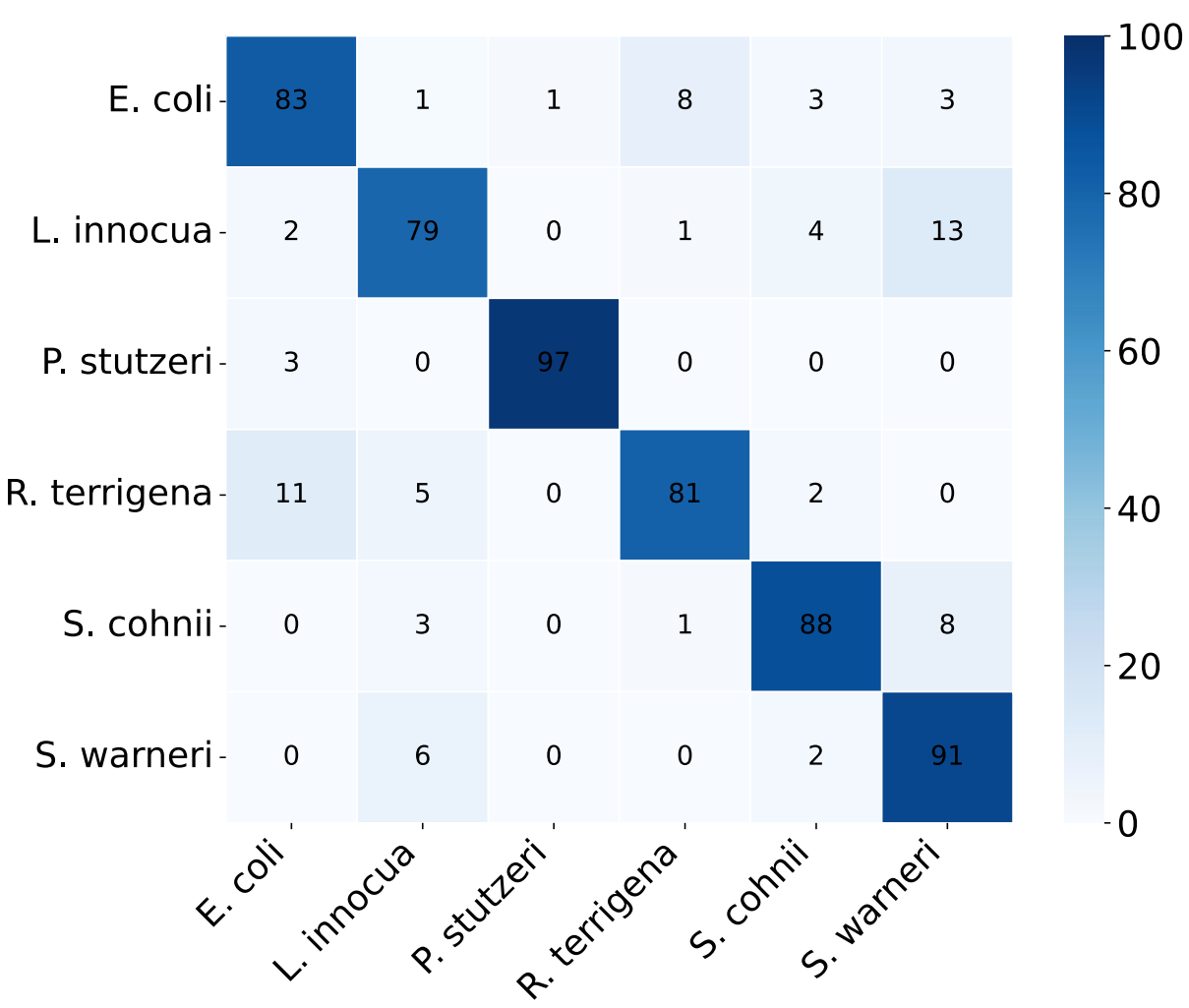

Fig. 6. Confusion matrix of the transformer classifier obtained from the combined predictions of all external test replicates.

Fig. 7 compares the feature spaces obtained by PCA, ICA, and the transformer model for Replicate 4, which corresponded to one of the test replicates where the transformer demonstrated a clear performance advantage over the conventional machine-learning approaches. For PCA and ICA, the score plots were generated using the components extracted by the respective feature-extraction methods. For the transformer model, the visualization was based on the final CLS-token embeddings extracted from the trained network before the classification layer. To enable visualization, the high-dimensional CLS-token embeddings were projected onto two principal components using PCA. These embeddings represent the learned latent feature space that the transformer uses to discriminate among bacterial species.

Clear differences in class separation can be observed among the three feature spaces. The PCA score plot (Fig. 7A) shows partial separation between bacterial species, although substantial overlap remains among several classes. The ICA representation (Fig. 7B) exhibits even stronger overlap, indicating limited discriminative power of the extracted components. In contrast, the transformer embeddings (Fig. 7C) form substantially more compact and well-separated clusters. Most bacterial species occupy distinct regions of the latent space, while only limited overlap remains between a small number of classes. The improved cluster separation is consistent with the superior classification performance observed for the transformer model on this replicate and suggests that the network learned more discriminative spectral representations than those obtained through conventional linear feature-extraction methods. Although the visualization is based on a two-dimensional PCA projection of the embedding space, the observed separation indicates that the transformer captured class-relevant information more effectively than the conventional approaches

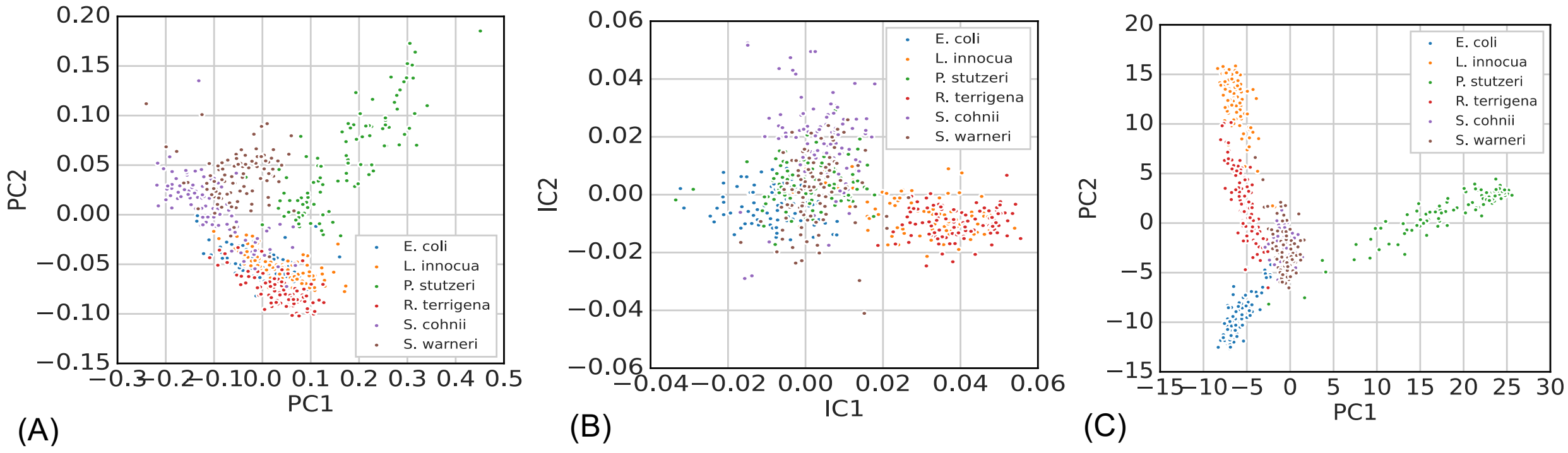

Fig. 7. Comparison of feature spaces obtained for Replicate 4 (external test replicate). (A) PCA score plot. (B) ICA score plot. (C) PCA score plot obtained from the final CLS-token embeddings.

### 3.2. Classification Performance Using Raw Raman Spectra

Classification performance was further assessed using the raw Raman spectra to examine the dependence of the investigated methods on spectral preprocessing. In this analysis, no baseline correction, noise removal, or vector normalization was performed prior to model training. All classification models were evaluated using the same nested leave-one-replicate-out cross-validation strategy described above.

Figure 8 summarizes the classification performance obtained from the raw Raman spectra. Overall, the same performance trends observed for the preprocessed data were largely preserved. The transformer model achieved the highest median sensitivity among all evaluated methods and maintained consistently strong performance across the nine independent measurement replicates (Fig. 8A). In contrast, the conventional machine-learning approaches showed greater variability between replicates, particularly for the Random Forest-based models.

The replicate-wise results shown in Fig. 8B further illustrate these differences. While PCA-SVM and ICA-SVM remained among the strongest conventional approaches and experienced only a moderate reduction in performance compared with the preprocessed data, larger fluctuations were observed across individual replicates. The transformer, in contrast, demonstrated more stable behavior, yielding consistently high sensitivity across the external validation folds. Notably, the classification performance of the transformer on raw spectra remained very close to that obtained using preprocessed spectra, indicating that the model was able to learn informative spectral representations directly from the original measurements.

Although preprocessing improved the performance of all evaluated methods, the transformer remained the best-performing approach and exhibited the most consistent behavior across independent measurement replicates. These findings demonstrate that transformer-based models can maintain robust classification performance even when applied directly to raw Raman spectra.

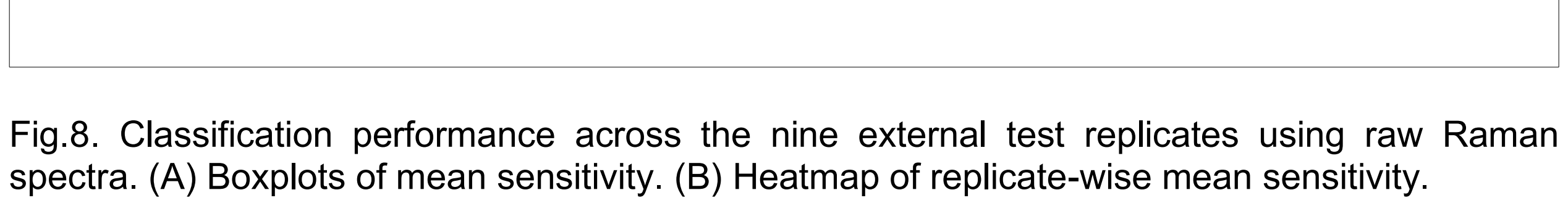

Fig.8. Classification performance across the nine external test replicates using raw Raman spectra. (A) Boxplots of mean sensitivity. (B) Heatmap of replicate-wise mean sensitivity.

## 4 Conclusion

This study systematically evaluated the performance of a transformer-based approach for bacterial Raman spectral classification using a nested leave-one-replicate-out cross-validation framework. The transformer model was compared with several conventional machine-learning pipelines combining PCA or ICA feature extraction with LDA, SVM, and Random Forest classifiers.

Across all independent measurement replicates, the transformer consistently achieved the highest classification performance and significantly outperformed the conventional approaches. The analysis of the learned feature space suggested that the transformer captured more discriminative latent representations than PCA and ICA, leading to improved class separation. These findings provide a potential explanation for the superior classification performance observed across the evaluation replicates.

The robustness of the transformer model was further demonstrated using raw Raman spectra. Although spectral preprocessing improved the performance of all investigated methods, the transformer maintained the highest classification performance and exhibited stable behavior across independent replicates of the measurement. These findings indicate that transformer-based models may reduce the dependence on extensive preprocessing compared with conventional machine-learning approaches.

Despite the superior classification performance observed in this study, transformer-based models remain more computationally demanding and less interpretable than conventional chemometric approaches. In contrast to PCA and ICA, where extracted components can often be directly related to spectral features, the latent representations learned by transformer models are more difficult to interpret.

**Funding**

This work is supported by the BMFTR, funding program Photonics Research Germany (FKZ: 13N15466 (LPI-BT1-FSU)), and is integrated into the Leibniz Center for Photonics in Infection Research (LPI). The LPI initiated by Leibniz-IPHT, Leibniz-HKI, UKJ, and FSU

Jena is part of the BMFTR national roadmap for research infrastructures. This work was co-funded by the European Union (European Research Council (ERC), STAIN-IT (Grant agreement No. 101088997), SpecReK (FKZ:13N16978).

## References

[1] T.W. Bocklitz, S. Guo, O. Ryabchykov, N. Vogler, J.r. Popp, Raman based molecular imaging and analytics: a magic bullet for biomedical applications!?, Analytical chemistry, 88 (2016) 133–151.

[2] K. Kong, C. Kendall, N. Stone, I. Notingher, Raman spectroscopy for medical diagnostics—From in-vitro biofluid assays to in-vivo cancer detection, Advanced drug delivery reviews, 89 (2015) 121–134.

[3] C.A.F. de Oliveira Penido, M.T.T. Pacheco, I.K. Lednev, L. Silveira Jr, Raman spectroscopy in forensic analysis: identification of cocaine and other illegal drugs of abuse, Journal of Raman Spectroscopy, 47 (2016) 28–38.

[4] B. Lorenz, C. Wichmann, S. Stöckel, P. Rösch, J. Popp, Cultivation-free Raman spectroscopic investigations of bacteria, Trends in microbiology, 25 (2017) 413–424.

[5] A. Paudel, D. Raijada, J. Rantanen, Raman spectroscopy in pharmaceutical product design, Advanced drug delivery reviews, 89 (2015) 3–20.

[6] W.J. Tipping, M. Lee, A. Serrels, V.G. Brunton, A.N. Hulme, Imaging drug uptake by bioorthogonal stimulated Raman scattering microscopy, Chemical Science, 8 (2017) 5606–5615.

[7] M. Shiota, M. Naya, T. Yamamoto, T. Hishiki, T. Tani, H. Takahashi, A. Kubo, D. Koike, M. Itoh, M. Ohmura, Gold-nanofève surface-enhanced Raman spectroscopy visualizes hypotaurine as a robust anti-oxidant consumed in cancer survival, Nature communications, 9 (2018) 1561.

[8] J. Desroches, M. Jermyn, M. Pinto, F. Picot, M.-A. Tremblay, S. Obaid, E. Marple, K. Urmey, D. Trudel, G. Soulez, A new method using Raman spectroscopy for in vivo targeted brain cancer tissue biopsy, Scientific reports, 8 (2018) 1792.

[9] T. Vankeirsbilck, A. Vercauteren, W. Baeyens, G. Van der Weken, F. Verpoort, G. Vergote, J.P. Remon, Applications of Raman spectroscopy in pharmaceutical analysis, TrAC trends in analytical chemistry, 21 (2002) 869–877.

[10]. P. Fu, Y. Wen, Y. Zhang, L. Li, Y. Feng, L. Yin, H. Yang, SpectraTr: A novel deep learning model for qualitative analysis of drug spectroscopy based on transformer structure, Journal of Innovative Optical Health Sciences, 15 (2022) 2250021.

[11] P. Lasch, Spectral pre-processing for biomedical vibrational spectroscopy and microspectroscopic imaging, Chemometrics and Intelligent Laboratory Systems, 117 (2012) 100–114.

[12] Å. Rinnan, F. Van Den Berg, S.B. Engelsen, Review of the most common pre-processing techniques for near-infrared spectra, TrAC Trends in Analytical Chemistry, 28 (2009) 1201–1222.

[13] H. Abdi, L.J. Williams, Principal component analysis, Wiley interdisciplinary reviews: computational statistics, 2 (2010) 433–459.

[14] R.G. Brereton, G.R. Lloyd, Support vector machines for classification and regression, Analyst, 135 (2010) 230–267.

[15] S. Guo, P. Rösch, J. Popp, T. Bocklitz, Modified PCA and PLS: Towards a better classification in Raman spectroscopy–based biological applications, Journal of Chemometrics, 34 (2020) e3202.

[16] C.-S. Ho, N. Jean, C.A. Hogan, L. Blackmon, S.S. Jeffrey, M. Holodniy, N. Banaei, A.A. Saleh, S. Ermon, J. Dionne, Rapid identification of pathogenic bacteria using Raman spectroscopy and deep learning, Nature communications, 10 (2019) 4927.

[17] J.M. Jafari, T. Bocklitz, Exploring feature extraction methods for Raman spectroscopy: A comparative study, Analytica Chimica Acta, (2025) 344755.

[18] O. Ryabchykov, S. Guo, T. Bocklitz, Analyzing Raman spectroscopic data, Physical Sciences Reviews, 4 (2019) 20170043.

[19] R. Luo, J. Popp, T. Bocklitz, Deep learning for Raman spectroscopy: A review, Analytica, 3 (2022) 287–301.

[20] J. Contreras, S. Mostafapour, J. Popp, T. Bocklitz, Siamese networks for clinically relevant bacteria classification based on raman spectroscopy, Molecules, 29 (2024) 1061.

[21] C. Chen, W. Wu, C. Chen, F. Chen, X. Dong, M. Ma, Z. Yan, X. Lv, Y. Ma, M. Zhu, Rapid diagnosis of lung cancer and glioma based on serum Raman spectroscopy combined with deep learning, Journal of Raman Spectroscopy, 52 (2021) 1798–1809.

[22]  Goodfellow, Y. Bengio, A. Courville, Y. Bengio, Deep learning, MIT press Cambridge2016.

[23] F. Lussier, V. Thibault, B. Charron, G.Q. Wallace, J.-F. Masson, Deep learning and artificial intelligence methods for Raman and surface-enhanced Raman scattering, TrAC Trends in Analytical Chemistry, 124 (2020) 115796.

[24] V. Ashish, Attention is all you need, Advances in neural information processing systems, 30 (2017) I.

[25] A. Dosovitskiy, L. Beyer, A. Kolesnikov, D. Weissenborn, X. Zhai, T. Unterthiner, M. Dehghani, M. Minderer, G. Heigold, S. Gelly, An image is worth 16x16 words: Transformers for image recognition at scale, arXiv preprint arXiv:2010.11929, (2020).

[26] K. Han, Y. Wang, H. Chen, X. Chen, J. Guo, Z. Liu, Y. Tang, A. Xiao, C. Xu, Y. Xu, A survey on vision transformer, IEEE transactions on pattern analysis and machine intelligence, 45 (2022) 87–110.

[27] P. Fu, Y. Wen, Y. Zhang, L. Li, Y. Feng, L. Yin, H. Yang, SpectraTr: A novel deep learning model for qualitative analysis of drug spectroscopy based on transformer structure, Journal of Innovative Optical Health Sciences, 15 (2022) 2250021.

[28] M. Chang, C. He, Y. Du, Y. Qiu, L. Wang, H. Chen, RaT: Raman Transformer for highly accurate melanoma detection with critical features visualization, Spectrochimica Acta Part A: Molecular and Biomolecular Spectroscopy, 305 (2024) 123475.

[29] Y. Zou, Y. Li, F. Zhang, Y. Ge, W. Wang, M. Chen, Rapid identification of Litopenaeus vannamei pathogenic bacteria: a combined approach using surface-enhanced Raman spectroscopy (SERS) and deep learning, Analytical and Bioanalytical Chemistry, 417 (2025) 4587–4603.

[30] H. Cao, J. Cheng, X. Ma, S. Liu, J. Guo, D. Li, Deep learning enabled open-set bacteria recognition using surface-enhanced Raman spectroscopy, Biosensors and Bioelectronics, 276 (2025) 117245.

[31] Y.-M. Tseng, K.-L. Chen, P.-H. Chao, Y.-Y. Han, N.-T. Huang, Deep learning–assisted surface-enhanced raman scattering for rapid bacterial identification, ACS Applied Materials & Interfaces, 15 (2023) 26398–26406.

[32] S. Guo, T. Bocklitz, U. Neugebauer, J. Popp, Common mistakes in cross-validating classification models, Analytical Methods, 9 (2017) 4410–4417.

[32] N. Ali, S. Girnus, P. Rösch, J.r. Popp, T. Bocklitz, Sample-size planning for multivariate data: a Raman-spectroscopy-based example, Analytical chemistry, 90 (2018) 12485–12492.